\documentclass[10pt,twocolumn,letterpaper]{article}

\usepackage{cvpr} 

\usepackage{cuted}
\usepackage{currfile}
\usepackage{caption}
\usepackage{microtype}
\usepackage{multirow}
\usepackage{array}
\usepackage{pifont}
\usepackage{mathtools}
\usepackage{bm}

\definecolor{cvprblue}{rgb}{0.21,0.49,0.74}
\usepackage[pagebackref,breaklinks,colorlinks,allcolors=cvprblue]{hyperref}

\def\paperID{14871}
\def\confName{CVPR}
\def\confYear{2027}

\newcommand{\method}{SearchChord}

\newcommand{\Rfield}{\mathcal{R}}

\title{\method: Prompt Preselection and Field Recomposition for One-Step Image Editing}
\author{Boyang Deng, Yuzhen Lu$^*$\\
Department of Biosystems and Agricultural Engineering, Michigan State University, USA\\
\ttfamily{luyuzhen@msu.edu}}

\renewcommand{\paragraph}[1]{\par\vspace{.5em}\noindent\textbf{#1}\enspace\ignorespaces}
\begin{document}
\maketitle

\begin{strip}
\centering
\includegraphics[width=.98\linewidth]{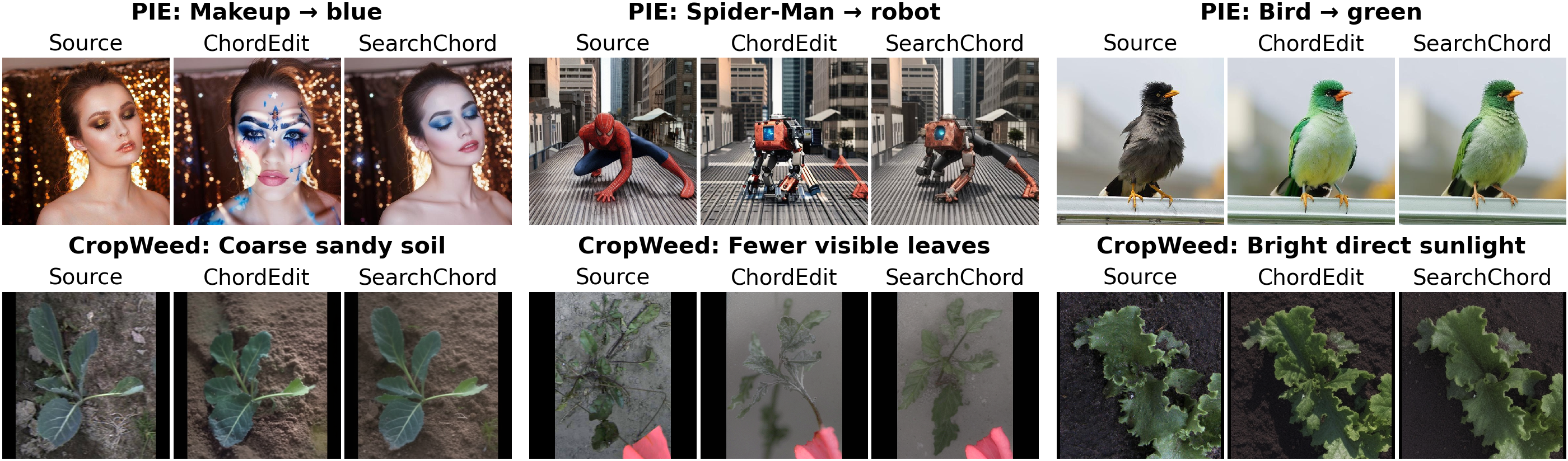}
\captionsetup{hypcap=false}
\captionof{figure}{\textbf{Edited examples: three PIE-Bench and three CropWeed examples.} Sources are paired with ChordEdit (one field) and SearchChord (TextSim $\rightarrow$ EditReward; 10 fields from 100 candidates), with common refinement.}
\captionsetup{hypcap=true}
\label{fig:comparison}
\end{strip}

\begin{abstract}
Finding effective prompts and seeds for one-step image editors requires balancing edit fidelity, source preservation, and computational cost. To address this challenge, we explore a large collection of candidate edits, where each candidate defined by a unique prompt–seed pair, and generate corresponding images on demand. We present \method, a training-free framework that leverages a large pool of such candidates, combining text-based preselection, image verification, and latent-field recomposition. Our ablation studies across pool sizes up to 200 candidates show that a pool of 100 offers a cost-effective operating point. Compared with feeding all 100 candidates into recomposition, TextSim preselects a subset of 10 candidates. Rather than uniform averaging, we incorporate image-verifier scores to weight each candidate during recomposition. Experiments on PIE-Bench and the proposed CropWeed-EditBench reveal a trade-off: candidate selection improves target-text alignment, while recomposition better preserves unmodified source content. The oracle verifier achieves the best performance in both preservation and target-text alignment. Practical verification partially recovers semantic alignment, but recomposition still compromises edit fidelity, including in agricultural image edits. These results motivate text-first preselection and stronger verification to balance preservation, fidelity, and cost. Code and data will be made publicly available.

\end{abstract}

\section{Introduction}
\label{sec:intro}

Text-guided image editing seeks to change requested attributes while preserving unrelated content. Recent one-step editors reduce the iterative cost of this task: SwiftEdit adapts one-step diffusion for fast editing \cite{nguyen2025swiftedit}, while ChordEdit uses training-free, inversion-free low-energy transport \cite{lu2026chordedit}. Editing-specific reward models, including EditReward \cite{wu2026editreward} and EditScore \cite{luo2026editscore}, complement these generators by evaluating generated edits. Together, these developments motivate studying how candidate generation and verification can balance preservation, edit fidelity, and computation.

Fast editors enable search over alternative prompts and random seeds, raising a central question: should we select one complete edit, combine candidate control fields, or preselect candidates before image generation? Stronger edits can disturb identity, geometry, and non-target content. Building on ChordEdit, the proposed \method{} ranks prompts with TextSim, our text-only score based on semantic similarity and directional alignment~\cite{reimers2019sentencebert,gal2022stylegannada}, and preselects candidates before computing fields, optionally verifies retained candidate images, and recomposes their fields before a common final generation step (Figure~\ref{fig:pipeline}). Shared candidate pools and fixed refinement isolate recomposition from changes to the base editor. Figure~\ref{fig:comparison} shows saved outputs.

Expanding the candidate pool offers more alternatives, but its benefit depends on candidate quality, selector reliability, and evaluation cost \cite{snell2025testtime,gao2023overoptimization,qu2026adaptiveediting}. This study investigates three questions: whether text-based preselection identifies useful candidates before generation; how field recomposition compares with selecting one candidate from the same pool; and how these choices affect preservation and target-instruction alignment. Strong preservation may reflect a weaker edit, while stronger transformations can disturb non-target content. This work therefore evaluates unchanged-region fidelity and target-text alignment separately on both benchmarks, using shared pools and fixed settings to assess trade-offs between quality and computation, and between preservation and semantic alignment.

The main contribution of this work is \textbf{\method{}}, a training-free framework for candidate search and field recomposition in one-step image editing. TextSim preselects 10 of 100 prompt/seed candidates before field computation; optional image verification then reweights the retained fields, which are combined before a common final generation step. The resulting pipeline integrates text-based candidate ranking, image-based assessment, and complete-field aggregation while retaining the base editor and refinement. With shared candidate pools, we compare single-candidate selection with equal- and score-weighted recomposition to isolate how these choices affect the final edit. Evaluations on PIE-Bench and CropWeed-EditBench, including separate illumination, soil, and relative leaf-amount edits for 3,387 plant instances, characterize the resulting trade-offs between preservation, target-text alignment, and computation, including the semantic loss that practical image verification only partially recovers.

\section{Related Work}
\label{sec:related}

\paragraph{Fast and training-free editing.}
Diffusion editing commonly uses inversion or iterative denoising to preserve source content \cite{mokady2023nulltext,ju2024pnp}. FlowEdit removes explicit inversion but integrates vector fields over multiple steps \cite{kulikov2025flowedit}. SwiftEdit and ChordEdit adapt editing to one-step generators \cite{nguyen2025swiftedit,lu2026chordedit}. A subsequent ChordEdit analysis examines the roles of timestep selection, transport, and proximal refinement \cite{li2026rethinking}. We retain the base estimator and refinement in the main search study; the matched-center controls isolate a separate estimator question.

\paragraph{Test-time scaling and verification.}
For LLMs, test-time scaling can improve outputs through additional inference compute: repeated sampling expands solution coverage, but imperfect verification can limit selection gains \cite{brown2024monkeys}; difficulty-aware search and revision improve budget use \cite{snell2025testtime}; Archon optimizes combinations of inference techniques \cite{saadfalcon2025archon}. These findings motivate candidate diversity, reliable verification, and adaptive compute allocation as promising directions for image editing. Diffusion search \cite{zhang2026diffusionsearch,aminian2026bon} similarly depends on reliable rewards \cite{gao2023overoptimization,yu2026pessimism}. ELECT, ADE-CoT, and GridAR use early filtering, adaptive budgets, and pruning \cite{kim2025elect,qu2026adaptiveediting,park2026gridar}. We combine fixed-budget text preselection, EditReward/EditScore verification \cite{wu2026editreward,luo2026editscore}, and complete Chord-field recomposition.

\paragraph{Field composition.}
SplitFlow decomposes multi-attribute target prompts and combines sub-target flows through projection and aggregation \cite{yoon2025splitflow}. We instead rank complete edit candidates for the same target and aggregate their Chord control fields before one final transport. This distinction concerns the candidate construction and composition unit, rather than the general idea of field aggregation. 

\paragraph{Agricultural generative vision.}
Agricultural imagery varies with illumination, soil type, plant growth stage, and season \cite{steininger2023cropandweed}. Generative augmentation can support plant recognition when outputs remain plausible and label-consistent \cite{deng2025weedaugmentation,tan2026agriculturaldiffusion}. The proposed CropWeed-EditBench evaluates appearance and relative plant-structure edits in this setting. Its image-editing metrics may help assess the suitability of edited images for downstream classification or detection tasks, such as those created by inserting plant instances into a field background.

\section{\method}
\label{sec:method}
\begin{figure*}[t]
\centering
\includegraphics[width=.98\textwidth]{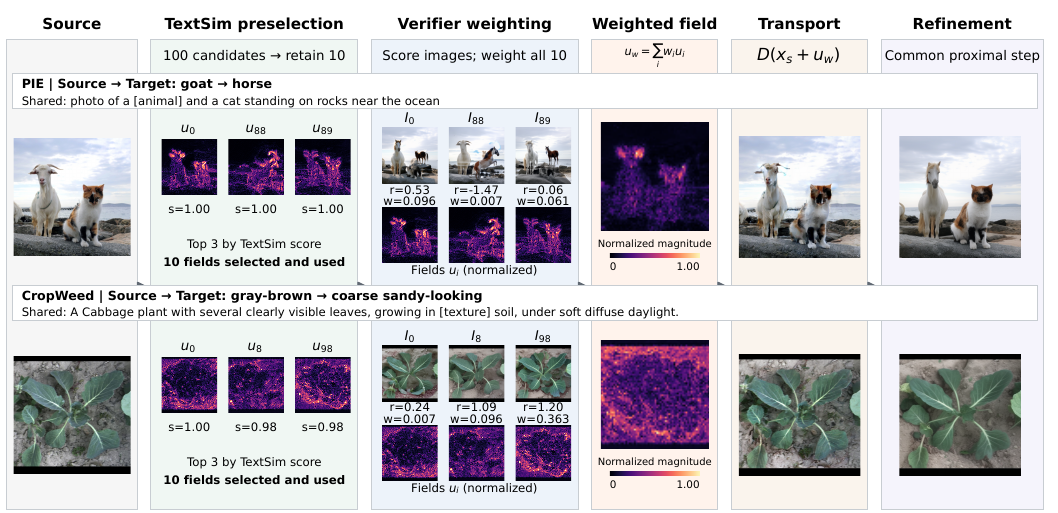}
\caption{\textbf{SearchChord pipeline.} The arrows indicate the words changed between the source and target prompts, while bracketed placeholders mark where those words appear within the shared sentence. }
\label{fig:pipeline}
\end{figure*}

\paragraph{Overview and notation.}
A \emph{candidate} pairs a target-prompt variant with a seed. Its \emph{control field} is a Chord latent update; applying the update, refining, and decoding produces a \emph{candidate image}. We consider $N$ candidates and combine $T$ fields into one final edit. Full-pool verification generates all $N$ candidate images before retaining $T$; TextSim ranks all $N$ texts and preselects $T$ candidates before computing their fields.

\subsection{Chord Candidate Fields}
The source image is denoted by $x$, its description by $c_s$, and the target prompt by $c_t$. Candidate $i$ uses prompt variant $\tilde c_{t,i}$ and seeded noise $\epsilon_i$. The residual $\Rfield_i(s)$ subtracts the source-conditioned clean-latent prediction from the target-conditioned prediction at the same noised source state and time $s$. It estimates the update associated with the requested change. Reusing candidate noise across times makes the temporal comparison use the same perturbation. The original Chord estimator combines two such residuals:
\begin{equation}
 u_i =
 \frac{t\,\Rfield_i(t-\delta)+\delta\,\Rfield_i(t)}
      {t+\delta},
\label{eq:chord}
\end{equation}
with the published default $t=0.90$, $\delta=0.15$, and transport scale $\lambda=1.0$ \cite{lu2026chordedit}. Unless otherwise stated, the main experiments also keep ChordEdit's proximal refinement fixed at $t_c=0.30$ \cite{lu2026chordedit}, so every selection strategy receives the same final semantic refinement.

\paragraph{Exploring alternative field estimators.}
ChordEdit uses causal one-sided kernel smoothing \cite{lu2026chordedit}; we propose BiChord to explore symmetric smoothing to reduce potential temporal bias.
The original asymmetric estimator is compared with \emph{BiChord}, which samples a symmetric temporal neighborhood,
\begin{equation}
 u_i^{\mathrm{Bi}}=
 \alpha\Rfield_i(t-\delta)
 +(1-2\alpha)\Rfield_i(t)
 +\alpha\Rfield_i(t+\delta),
 \label{eq:bichord}
\end{equation}
with $\alpha=0.25$. Its weights sum to one, $\alpha+(1-2\alpha)+\alpha=1$, so no denominator is needed, whereas Chord normalizes by $t+\delta$. BiChord uses $t=0.80$, $\delta=0.15$, and timesteps $(0.65,0.80,0.95)$ within the diffusion interval. A Chord control at $t=0.80$ isolates the estimator from the center-time change.

\emph{BiChord-RA} makes the side weight risk-adaptive rather than fixed. Let $\sigma_i^2$ denote a local residual-variability estimate and $C_i$ a local curvature estimate around $t$. We use
\begin{equation}
 \alpha_i^{\mathrm{RA}}=
 \operatorname{clip}_{[0,\,1/2]}
 \left(\frac{2\sigma_i^2}{\delta^4 C_i^2+6\sigma_i^2}\right).
 \label{eq:bichordra}
\end{equation}
For each image, $\sigma_i^2$ averages the population residual variance across two noise draws and the three diffusion timesteps. The curvature term $C_i^2$ is the mean squared second difference of the noise-averaged residuals divided by $\delta^4$, over latent channels and positions. Greater curvature reduces the side weight in Eq.~\ref{eq:bichordra}, keeping the estimate closer to the center residual. Table~\ref{tab:repro} shares IDs, two noise samples, and metric denominators across all four estimators.

\subsection{Candidate Construction}
This study explores candidate budgets from $N=1$ to $N=200$, using $N=100$ as baseline experiments to balance diversity and computation. Each candidate pairs a prompt variant with a noise seed. FLAN-T5-base~\cite{chung2024flan} generates distinct prompt variants to preserve the target meaning; we deduplicate each batch and fill shortfalls with the original target using fresh seeds. Larger pools retain earlier candidates for controlled comparisons, with their average composition shown in Figure~\ref{fig:scaling}(b). Because textually similar prompt variants may alter meaning, final refinement uses the original target prompt to reinforce the requested edit.

\subsection{TextSim Preselection}
TextSim assesses each prompt variant before image generation using cosine similarity between text embeddings~\cite{reimers2019sentencebert} and a text-only adaptation of directional alignment~\cite{gal2022stylegannada}. We average the generator text encoder's token features using the attention mask to exclude padding, and apply L2 normalization to the resulting representations. Let $e_s=f(c_s)$, $e_t=f(c_t)$, and $e_i=f(\tilde c_{t,i})$ be the resulting source, target, and candidate vectors. The score combines two forms of alignment:
\begin{align}
 s_i^{\mathrm{tar}} &= \cos(e_i,e_t),\\
 s_i^{\mathrm{dir}} &= \cos(e_i-e_s,e_t-e_s),\\
 s_i^{\mathrm{text}} &= \alpha s_i^{\mathrm{tar}}
 +(1-\alpha)s_i^{\mathrm{dir}},\qquad \alpha=0.5.
\label{eq:textsim}
\end{align}
Target alignment measures how closely a prompt variant matches the target description, while direction alignment measures whether the change from the source description aligns with the requested edit. We retain the $T$ highest-scoring candidates, compute their fields, and combine them using text-derived weights, as described below. For example, $(N,T)=(100,10)$ ranks 100 candidates but computes only 10 fields. Notably, TextSim operates solely on text: it cannot detect visual artifacts, distinguish different seeds paired with the same prompt, or guarantee that prompt variants preserve the intended meaning. Despite these limitations, it is hypothesized that preselecting 10 candidates from a pool of 100 and averaging their control fields can improve editing. We assess this potential through separate measures of source preservation and edit fidelity.

\subsection{Text-First Verification Pipeline}
A two-stage verification pipeline is used to combine text-based preselection with image-based assessment. TextSim first retains 10 candidates from a pool of 100. We compute their control fields and generate the corresponding candidate images. A verifier then evaluates each of these 10 edits using the source image, edit instruction, and generated output. Our experiments use EditReward and EditScore, but the same procedure can accommodate other compatible image verifiers. The two stages are summarized as
\begin{equation}
 \begin{aligned}
 100\text{ candidates}&\xrightarrow{\text{TextSim}}10\text{ fields},\\
 10\text{ candidate images}&\xrightarrow{\text{Verifier}}10\text{ scores}.
 \end{aligned}
 \label{eq:cascade}
\end{equation}
For field recomposition, the verifier scores determine the weights assigned to the same 10 fields. All 10 candidates could contribute to the combined field (Weighted Top-T), or instead returns the highest-scoring candidate (Best-of-T), enabling the assessment of whether image-based verification improves on text-only weighting. A verifier can detect visual artifacts in preselected candidates that TextSim cannot assess. Figure~\ref{fig:pipeline} illustrates the pipeline and the resulting transport and refinement outputs.

\subsection{Practical Verification and Diagnostic Oracle}
Practical verifiers assess each candidate against the source image and the original editing instruction. Let $x$ denote the source image, $d$ the instruction, and $I_i$ a candidate output. The verifier score is
\begin{equation}
 r_i=\mathcal{V}(x,d,I_i),
 \label{eq:verifier_score}
\end{equation}
where $\mathcal{V}$ is EditReward or EditScore~\cite{wu2026editreward,luo2026editscore}. The instruction $d$ specifies a transformation, such as ``replace the goat with a horse''. These verifiers estimate edit quality and instruction following without benchmark masks.

We use \emph{oracle} for a verifier with access to privileged evaluation information, following the oracle-verifier usage in inference-time sampling~\cite{brown2024monkeys}. Our oracle uses benchmark masks and directly computed evaluation metrics to rank the candidate pool. It provides a diagnostic reference under the chosen metrics.

Preservation and semantics each receive 50\% total weight, which avoids letting preservation dominate simply because it has three reported metrics:
\begin{equation}
\begin{split}
S_i^{(5)}={}&\tfrac16 r_i^{\mathrm{PSNR}}
+\tfrac16 r_i^{\mathrm{MSE}}
+\tfrac16 r_i^{\mathrm{LPIPS}}\\
&+\tfrac14 r_i^{\mathrm{CLIP-W}}
+\tfrac14 r_i^{\mathrm{CLIP-E}}.
\end{split}
\label{eq:oracle5}
\end{equation}
Here $r_i^{m}$ denotes the normalized rank of candidate $i$ under metric $m$ among the same $N$ candidates, where m is a placeholder for the five metrics (PSNR, MSE, LPIPS, CLIP-W, CLIP-E). Higher PSNR and CLIP scores receive higher ranks, whereas lower MSE and LPIPS errors receive higher ranks. For each metric, candidates are ordered from worst to best. A candidate at position $k$ receives the normalized rank $(k-1)/(N-1)$, giving 0 to the worst and 1 to the best when there are no ties. Candidates with tied metric values share the average normalized rank of their positions. Rank normalization makes the metric units comparable but reflects only relative standing within the candidate pool. 

\subsection{\texorpdfstring{Top-$T$}{Top-T} Field Recomposition}
Three ways are compared to use the candidate fields: selecting the highest-scoring candidate (Best-of-$N$), averaging the retained fields, and combining them with score-based weights. Let $\mathcal{T}_{N,T}$ denote the $T$ candidates retained from a pool of $N$.  Best-of-$N$ uses a single field ($T=1$), while mean and weighted recomposition combine the retained fields as
\begin{align}
 u_{\mathrm{mean}} &= \frac{1}{T}\sum_{i\in\mathcal{T}_{N,T}}u_i,\\
 u_{\mathrm{w}} &= \sum_{i\in\mathcal{T}_{N,T}}w_i u_i,\notag\\
 w_i &= \frac{\exp(\tilde S_i/\tau)}
 {\sum_{j\in\mathcal{T}_{N,T}}\exp(\tilde S_j/\tau)}.
\label{eq:aggregation}
\end{align}
For TextSim and practical verifiers, $\tilde S_i$ is the normalized score rank; for the oracle, it is the balanced score in Eq.~\ref{eq:oracle5}. Temperatures $\tau=0.25$ is used to balance emphasis on top candidates with weight diversity.

To express the resulting update relative to the baseline, the final velocity latent update is:
\begin{equation}
 u_\beta=u_0+\beta(u_{\mathrm{agg}}-u_0),
\label{eq:anchor}
\end{equation}
where $u_0$ is the field obtained with the original target prompt and base seed, and $u_{\mathrm{agg}}$ is either $u_{\mathrm{mean}}$ or $u_{\mathrm{w}}$. We set $\beta=1$, as $\beta=0.5$ or $\beta=2$ performed worse in preliminary experiments.

The combined update is applied to the encoded source latent $z$ through $z'=z+\lambda u_\beta$. We then apply the same proximal refinement with the original target prompt and base seed before decoding the final image, isolating the refinement settings from the effect of field combination.

\section{Experiments}
\label{sec:experiments}
\suppressfloats[t] 

\subsection{Benchmarks and Evaluation Protocol}
\label{sec:protocol}
\subsubsection{PIE-Bench}
The 700-sample PIE-Bench benchmark is used for direct comparison with prior real-image editing work \cite{ju2024pnp,lu2026chordedit}. The first candidate is the original target prompt with seed 42. ChordEdit baseline (one candidate) is compared with 100-candidate-based SearchChord variants with aggregation sizes $T=1$, $10$, and $100$. 

We use the unchanged PIE evaluator with CLIP ViT-L/14 and LPIPS SqueezeNet. For PSNR, MSE, and LPIPS, the edited area is zeroed in both source and output before full-frame evaluation. CLIP-Whole uses the full output; CLIP-Edited zeros pixels outside the annotated edit region. Both CLIP metrics are evaluated on all 700 shared samples. PSNR, MSE, and LPIPS use the same 556 samples whose unedited fraction is at least $0.001$. The same evaluation subsets are used across all PIE-Bench experiments. Masks are used for evaluation, and they do not constrain generation process.

\begin{table}[t]
\centering
\small
\caption{ChordEdit-based estimator comparison on PIE-Bench. Bold marks the best value in each metric column.}
\label{tab:repro}
\begin{tabular}{lccc}
\toprule
Field estimator & $t$ & PSNR $\uparrow$ & CLIP-Edited $\uparrow$\\
\midrule
Chord (default) & .90 & 23.52 & \textbf{21.85}\\
Chord (matched center) & .80 & \textbf{24.37} & 21.66\\
BiChord ($\alpha=.25$) & .80 & 23.71 & 21.67\\
BiChord-RA & .80 & 23.53 & 21.72\\
\bottomrule
\end{tabular}
\end{table}

\begin{table*}[t]
\centering
\scriptsize
\caption{Candidate scoring methods and information access. ``Image?'' indicates whether the edited candidate must already be generated. ``Benchmark GT/mask?'' indicates whether the method uses mask information for evaluation. ``Deployable?'' indicates whether the method can be used in a practical deployment setting. }
\label{tab:selector_hierarchy}
\resizebox{\textwidth}{!}{%
\begin{tabular}{lcccccc}
\toprule
Scoring method & Stage & Image? & Benchmark GT/mask? & Deployable? & Relative scoring cost & Main role\\
\midrule
Joint-5 balanced & diagnostic & Yes & Yes & No & Very high & primary headroom estimate\\
TextSim & pre-generation & No & No & Yes & Very low & reduce field/image budget\\
EditReward & post-generation & Yes & No & Yes & High & edit quality + instruction following\\
EditScore & post-generation & Yes & No & Yes & High & visual quality + instruction following\\
\bottomrule
\end{tabular}}
\end{table*}

\begin{table*}[t]
\centering
\scriptsize
\definecolor{tableverifierbest}{HTML}{245C96}
\definecolor{tabletextbest}{HTML}{006D5B}
\caption{PIE-Bench quality--compute comparison with common proximal refinement and $\beta=1$. All rows below the ChordEdit baseline are \method{} variants. Estimated total includes prompt-pool construction. Black bold marks overall column-best values; \textcolor{tableverifierbest}{blue bold} and \textcolor{tabletextbest}{teal bold} mark the best within the full-verifier (EditReward/EditScore) and TextSim/cascade blocks, respectively.}
\label{tab:main}
\setlength{\tabcolsep}{2.1pt}
\resizebox{\textwidth}{!}{%
\begin{tabular}{lccc|ccc|cc|cc}
\toprule
\multirow{2}{*}{Method / selector} &
\multirow{2}{*}{$N$} &
\multirow{2}{*}{$T$} &
\multirow{2}{*}{Fields} &
\multicolumn{3}{c|}{Unedited region} &
\multicolumn{2}{c|}{CLIP Semantics} &
\multicolumn{2}{c}{Time (s)}\\
\cmidrule(lr){5-7}\cmidrule(lr){8-9}\cmidrule(lr){10-11}
& & & &
PSNR $\uparrow$ &
MSE$\times10^3$ $\downarrow$ &
LPIPS$\times10^3$ $\downarrow$ &
Whole $\uparrow$ &
Edited $\uparrow$ &
Editing/scoring $\downarrow$ & Est. total $\downarrow$\\
\midrule
ChordEdit baseline & 1 & 1 & 1 & 22.44 & 8.28 & 121.06 & 25.87 & 22.58 & \textbf{0.49} & \textbf{0.49}\\
\midrule
Joint-5 oracle: Best-of-$N$$^\dagger$ & 100 & 1 & 100 & 23.39 & 6.95 & 110.58 & \textbf{27.93} & \textbf{25.22} & 146.55 & 175.63\\
Joint-5 oracle: Mean Top-$T$$^\dagger$ & 100 & 10 & 100 & 24.72 & 5.08 & 91.02 & 24.52 & 21.13 & 146.98 & 176.06\\
Joint-5 oracle: Weighted Top-$T$$^\dagger$ & 100 & 10 & 100 & \textbf{24.73} & \textbf{5.07} & \textbf{91.00} & 24.52 & 21.17 & 146.74 & 175.83\\
Joint-5 oracle: Mean Top-$T$$^\dagger$ & 100 & 100 & 100 & 24.53 & 5.21 & 91.40 & 24.13 & 20.83 & 146.74 & 175.83\\
Joint-5 oracle: Weighted Top-$T$$^\dagger$ & 100 & 100 & 100 & 24.69 & 5.08 & \textbf{91.00} & 24.25 & 20.93 & 146.74 & 175.83\\
\midrule
EditReward: Best-of-$N$ & 100 & 1 & 100 & 21.48 & 10.07 & 134.76 & \textcolor{tableverifierbest}{\textbf{26.32}} & \textcolor{tableverifierbest}{\textbf{23.08}} & 135.60 & 164.68\\
EditReward: Weighted Top-$T$ & 100 & 10 & 100 & 24.20 & 5.56 & 93.22 & 24.56 & 21.16 & 135.79 & 164.88\\
EditReward: Weighted Top-$T$ & 100 & 100 & 100 & 24.46 & 5.29 & 91.92 & 24.32 & 20.95 & 135.79 & 164.88\\
EditScore: Best-of-$N$ & 100 & 1 & 100 & 21.98 & 9.18 & 126.01 & 25.77 & 22.60 & \textcolor{tableverifierbest}{\textbf{109.46}} & \textcolor{tableverifierbest}{\textbf{138.55}}\\
EditScore: Weighted Top-$T$ & 100 & 10 & 100 & 24.34 & 5.43 & 92.39 & 24.41 & 21.08 & 109.67 & 138.75\\
EditScore: Weighted Top-$T$ & 100 & 100 & 100 & \textcolor{tableverifierbest}{\textbf{24.49}} & \textcolor{tableverifierbest}{\textbf{5.25}} & \textcolor{tableverifierbest}{\textbf{91.65}} & 24.26 & 20.92 & 109.66 & 138.75\\
\midrule
TextSim: Best-of-$N$ & 100 & 1 & 1 & 22.44 & 8.28 & 121.06 & 25.87 & 22.58 & \textcolor{tabletextbest}{\textbf{0.55}} & \textcolor{tabletextbest}{\textbf{29.64}}\\
TextSim: Weighted Top-$T$ & 100 & 10 & 10 & \textcolor{tabletextbest}{\textbf{24.39}} & \textcolor{tabletextbest}{\textbf{5.41}} & \textcolor{tabletextbest}{\textbf{92.67}} & 24.48 & 21.13 & 2.54 & 31.63\\
TextSim $\rightarrow$ EditReward: Best-of-10 & 100 & 1 & 10 & 22.13 & 9.00 & 126.92 & \textcolor{tabletextbest}{\textbf{26.49}} & \textcolor{tabletextbest}{\textbf{23.14}} & 13.80 & 42.89\\
TextSim $\rightarrow$ EditReward: Weighted & 100 & 10 & 10 & 24.13 & 5.70 & 94.46 & 24.72 & 21.39 & 14.01 & 43.09\\
TextSim $\rightarrow$ EditScore: Best-of-10 & 100 & 1 & 10 & 22.30 & 8.63 & 124.37 & 25.98 & 22.63 & 11.02 & 40.11\\
TextSim $\rightarrow$ EditScore: Weighted & 100 & 10 & 10 & 24.17 & 5.64 & 94.10 & 24.63 & 21.33 & 11.23 & 40.32\\
\bottomrule
\end{tabular}}
\vspace{0.25em}

{\scriptsize $^\dagger$Diagnostic oracle. Estimated total adds the separately measured prompt-pool construction mean for the same images: 29.09\,s per $N=100$ edit (FLAN-T5-base, FP32 CPU, eight threads). ChordEdit ($N=1$) requires no prompt-pool construction.}

\end{table*}

\begin{figure*}[t]
\centering
\includegraphics[width=.94\textwidth]{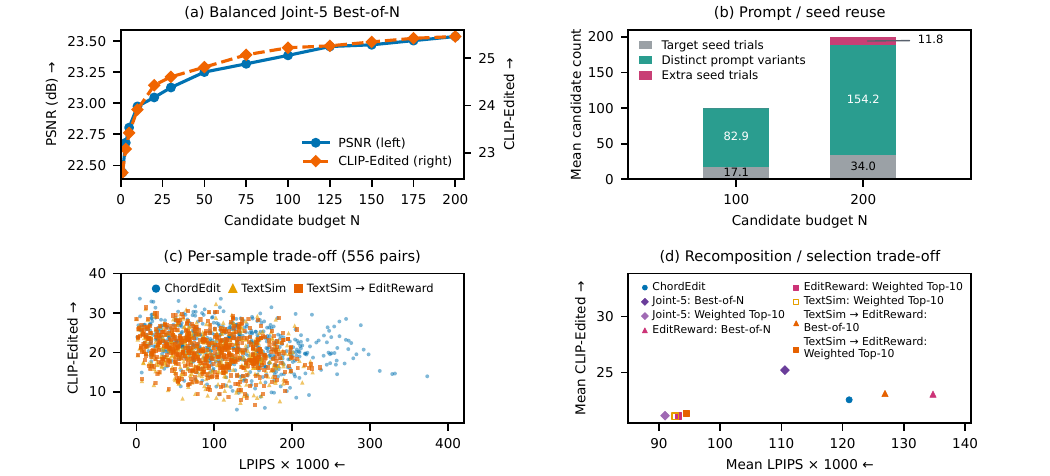}
\captionsetup{skip=4pt}
\caption{\textbf{Effects of candidate budget and field recomposition.} (a) PSNR and CLIP-Edited for balanced Joint-5 Best-of-$N$ as the candidate pool grows. (b) Pool composition averaged over 700 edits. Gray counts trials using the original target prompt, teal counts the first trial of each distinct prompt variant, and rose counts further trials of those variants with new seeds. (c) Per-image LPIPS and CLIP-Edited scores for methods evaluated on the same images. (d) Mean scores for candidate selection and field recomposition using $N=100$, with ChordEdit shown as a reference.}
\label{fig:scaling}
\end{figure*}

\begin{table*}[t]
\centering
\footnotesize
\caption{Separate illumination, soil, and relative leaf-amount evaluations for 3,387 CropWeed-EditBench plant instances. Bold marks the best values within each group.}
\label{tab:cropweed_results}
\setlength{\tabcolsep}{4.8pt}
\begin{tabular}{llccccc}
\toprule
& & \multicolumn{3}{c}{Unedited region} & \multicolumn{2}{c}{CLIP Semantics}\\
\cmidrule(lr){3-5}\cmidrule(lr){6-7}
Edit group & Method & PSNR $\uparrow$ & MSE$\times10^3$ $\downarrow$ & LPIPS$\times10^3$ $\downarrow$ & Whole $\uparrow$ & Edited $\uparrow$\\
\midrule
\multirow{3}{*}{Illumination/weather}
 & ChordEdit baseline & 25.87 & 3.07 & 95.87 & \textbf{25.39} & \textbf{20.39}\\
 & TextSim: Weighted Top-$T$ & \textbf{26.53} & \textbf{2.65} & 93.23 & 25.27 & 19.50\\
 & TextSim $\rightarrow$ EditReward: Weighted & 26.47 & 2.68 & \textbf{93.07} & 25.32 & 19.58\\
\midrule
\multirow{3}{*}{Soil appearance/type}
 & ChordEdit baseline & 25.24 & 3.62 & 103.71 & \textbf{26.52} & \textbf{22.13}\\
 & TextSim: Weighted Top-$T$ & \textbf{25.85} & \textbf{3.16} & 97.41 & 26.44 & 20.59\\
 & TextSim $\rightarrow$ EditReward: Weighted & 25.82 & 3.18 & \textbf{97.31} & 26.46 & 20.78\\
\midrule
\multirow{3}{*}{Relative leaf amount}
 & ChordEdit baseline & 25.39 & 3.56 & 119.52 & 25.32 & \textbf{24.54}\\
 & TextSim: Weighted Top-$T$ & \textbf{25.50} & \textbf{3.52} & 113.51 & 25.49 & 24.45\\
 & TextSim $\rightarrow$ EditReward: Weighted & 25.45 & 3.55 & \textbf{112.44} & \textbf{25.53} & 24.51\\
\bottomrule
\end{tabular}
\end{table*}

\subsubsection{CropWeed-EditBench}
The proposed CropWeed-EditBench dataset consists of 3,387 crop/weed bounding-box instances from top-down field images. Each plant instance includes a 10\% surrounding context margin with at least $300\times300$ pixels. Each image has three PIE-compatible source/target prompt pairs, including illumination, soil type, and relative leaf amount, sharing one source prompt. We use Qwen3-VL-2B-Instruct~\cite{bai2025qwen3vl}, run locally with 4-bit quantization, to infer attributes from each source image. These attributes populate fixed prompt templates. Each target prompt only changes the attribute to be edited, while keeping the rest of the source description unchanged.

The three categories are evaluated separately across the 3,387 plant instances, including illumination, soil-type, and relative leaf-amount specifications. All methods use the same source images, prompts, and masks. 

\paragraph{Image preparation and mask checks.}
CropWeed-EditBench masks are generated using SAM2.1 (Hiera-base-plus)~\cite{ravi2024sam2}. Images are resized to a maximum side of 768 pixels while preserving aspect ratio. We identify We threshold RGB and HSV channels to identify leafy vegetation pixels, which we feed as foreground point prompts. SAM2.1 then outputs candidate masks, ranked by predicted mask quality, vegetation coverage, and color agreement with green vegetation; the top-ranked mask is retained.

Leaf edits are evaluated within the plant region, while illumination and soil edits use a shared background mask. Notably, Leaf prompts request more or fewer leaves; CLIP-Edited does not establish exact leaf counts or biological validity. The masks are generated automatically, and only low-confidence predictions are flagged for review and correction. Each mask is resized to the original image dimensions using nearest-neighbor interpolation. Each plant image and its mask are then centered on a square canvas with black padding and resized to $512\times512$ pixels, using Lanczos interpolation for RGB images and nearest-neighbor interpolation for binary masks. This preserves the aspect ratio and retains the full image. Masks are stored as PIE-compatible, zero-based run-length pairs in row-major order. For every image, encoding and decoding recover the same mask, and the mask remains aligned with the image. We visually inspect representative image-mask overlays. 

\subsection{Main PIE-Bench Comparison}
\paragraph{ChordEdit-based estimators.}
Table~\ref{tab:repro} compares Chord, a matched-center Chord control, BiChord, and BiChord-RA. All four one-edit-step configurations use two noise samples, seed 42, $\delta=0.15$, and proximal refinement at $t_c=0.30$. The Chord control at $t=0.80$ matches the center timestep of BiChord and BiChord-RA. These two-noise configurations are evaluated separately from the one-noise baseline used in the main experiments. Chord attains the highest PSNR at $t=0.80$ and the highest CLIP-Edited at $t=0.90$. At the matched center, BiChord and BiChord-RA yield only small CLIP-Edited gains over Chord while reducing PSNR, so neither improves both metrics. A possible explanation is that residuals at different noise levels may point in different editing directions, allowing symmetric averaging to weaken useful components of the update. The adaptive rule responds to residual variability and curvature, which do not directly measure preservation or target alignment. These results suggest that symmetric smoothing alone is insufficient to improve editing quality. We therefore retain the original ChordEdit estimator with its published default $t=0.90$ for subsequent experiments.

\paragraph{Candidate scoring methods.}
Table~\ref{tab:selector_hierarchy} summarizes how candidate scoring methods differ in their roles and information requirements. The balanced Joint-5 oracle scores candidate images using benchmark metrics and masks to assess the quality available in the candidate pool. TextSim ranks prompts before image generation, requiring neither candidate images nor benchmark masks. EditReward and EditScore assess generated edits for quality and instruction following without benchmark masks, but require image generation and verifier computation. This distinction motivates our two-stage pipeline: TextSim limits the number of candidates passed to image verification, while the second verifier provides a further score for the candidate quality.

\paragraph{Selection versus recomposition.}
Table~\ref{tab:main} compares ChordEdit with variants of \method{}. 
Oracle Best-of-$100$ outperforms ChordEdit baseline: PSNR rises from 22.44 to 23.39 dB and CLIP-Edited from 22.58 to 25.22. Weighted Top-10 further raises PSNR to 24.73 and lowers LPIPS to 91.00, but reduces CLIP-Edited to 21.17. Field recomposition generally preserves unchanged regions better, while candidate selection (Best-of-N) achieves stronger alignment with the target prompt. Weighted recomposition generally performs slightly better than simple averaging, and combining the top 10 candidates gives better results than using all 100. 

\paragraph{Practical selection and recomposition.}
Practical verifiers show a similar trade-off between target alignment and preservation. With EditReward, Best-of-100 achieves a higher CLIP-Edited score than weighted Top-10 recomposition (23.08 versus 21.16), but a lower PSNR (21.48 versus 24.20 dB). EditScore shows the same trade-off, with slightly lower CLIP-Edited scores but slightly higher PSNR than EditReward. Both Best-of-100 methods remain below the diagnostic Joint-5 oracle's CLIP-Edited score of 25.22.

TextSim Best-of-100 selects the first candidate for all 700 images and therefore reproduces the baseline results. The original target prompt receives the highest text similarity score, but TextSim cannot distinguish candidates that use the same prompt with different seeds; the tie-breaking rule selects the first candidate. Combining the TextSim Top-10 fields instead raises PSNR to 24.39 dB, while lowering CLIP-Edited to 21.13.

The same trade-off remains after TextSim preselects 10 candidates. For both EditReward and EditScore, Best-of-10 achieves higher CLIP-Edited scores, while weighted recomposition gives higher PSNR. Image verification thus improves target alignment when selecting a single candidate, while combining fields better preserves unchanged regions. 

\subsection{Candidate Budgets and Quality Trade-offs}
\label{sec:budget}

Figure~\ref{fig:scaling}(a) shows that a larger candidate pool improves the quality available to the oracle. Increasing $N$ from 100 to 200 raises Joint-5 Best-of-$N$ PSNR from 23.39 to 23.54 dB and CLIP-Edited from 25.22 to 25.46.

Figure~\ref{fig:scaling}(b) shows how these pools combine prompt and seed variation. At $N=100$, each pool averages 17.1 candidates of the original target prompt and 82.9 distinct prompt variants. At $N=200$, the averages are 34.0 original-prompt candidates, 154.2 distinct variants, and 11.8 additional seed candidates reusing those variants. Thus, increasing $N$ adds both new prompts and further candidates of existing prompts.

In addition, in a small supplementary experiment, it is found that at matched per-image candidate counts, varying seeds with the original target prompt yielded results comparable to varying both prompts and seeds. Because the two groups use different seeds, this diagnostic comparison does not isolate the effect of prompt changes. Future work will use shared seeds and controlled candidate budgets to examine their separate contributions and interactions.

Panel (c) shows overlapping distributions on the same preservation-valid images. Panel (d) compares mean results at $N=100$: oracle selection favors semantics, while recomposition favors preservation. Neither dominates every metric; prefix ranks can also yield non-monotonic metric trends.

\paragraph{Compute accounting.}
Table~\ref{tab:main} shows that TextSim preselection reduces editing/scoring time relative to full-pool image verification. Prompt construction dominates its estimated total time, and ChordEdit remains faster overall. Totals sum separately measured CPU preparation and GPU execution on one RTX A6000, excluding model loading, I/O, and final evaluation. Generator-only peak allocation was 5.77\,GiB (FP32, text batch 128), excluding verifiers and allocator/driver overhead. Fewer fields reduce computation, while deployed memory also depends on the used generator and verifier.

\subsection{CropWeed-EditBench Evaluation}
We compare ChordEdit ($N=T=1$), TextSim Weighted Top-10 ($N=100$, $T=10$), and TextSim $\rightarrow$ EditReward Weighted Top-10 ($N=100$, $T=10$), following Section~\ref{sec:protocol}. All retain the ChordEdit settings, with one noise sample, base seed 42, and $\beta=1$, without edit-family-specific tuning. The two search variants share prompt/seed pools and retained candidate indices.

\paragraph{Preservation and edit fidelity.}
Table~\ref{tab:cropweed_results} shows the same preservation--alignment trade-off observed on PIE-Bench. TextSim improves mean PSNR over ChordEdit in all three edit categories, with larger gains for illumination and soil than for leaf amount, but lowers CLIP-Edited. EditReward weighting reduces LPIPS in every category and partially recovers target alignment relative to TextSim, with small decreases in PSNR. Its CLIP-Edited scores nevertheless remain below ChordEdit. Figure~\ref{fig:comparison} provides selected examples of soil, lighting, and foliage changes. These results demonstrate improved preservation, while leaving room to improve the requested edits.

\paragraph{What the current metrics capture.}
For illumination and soil edits, preservation scores measure how well the plant is retained; for leaf edits, they measure preservation outside the plant. Scores should therefore be compared within each edit category; black padding also contributes to preservation scores. The visual examples (Figure~\ref{fig:comparison}) show more convincing soil changes, while illumination and leaf edits remain less successful in producing the requested changes. These limitations highlight that strong preservation scores do not necessarily indicate successful edits. 

\paragraph{Automatic evaluation.}
An important next step is to complement CLIP-Edited with an automatic check of leaf-count changes. A pretrained counter such as CountGD~\cite{amininaieni2024countgd}, or a zero-shot leaf-instance segmentation method such as Leaf Only SAM~\cite{williams2023leafonlysam}, could estimate visible leaves on the focal plant before and after editing. We could then report how often the estimated leaf count increases when more leaves are requested or decreases when fewer leaves are requested, without manually counting leaves in every edited image. This metric has not been evaluated in this study; its reliability under leaf overlap, species differences, and generated artifacts would need validation.

\section{Conclusion}
This study proposed \method{}, a training-free framework for one-step image editing that preselects candidate prompts, optionally verifies the resulting edits, and recomposes retained fields. Results across both benchmarks show that selecting a single candidate improves target-text alignment, whereas combining multiple fields better preserves unchanged content. Although larger candidate pools improve oracle performance, a gap remains under practical verification. TextSim reduces editing and scoring costs compared with full-pool verification, albeit with additional prompt preparation overhead.

We also introduced CropWeed-EditBench, a benchmark for agricultural image editing covering illumination, soil, and relative leaf-amount modifications through controlled editing instructions and shared evaluation regions. Future work will explore edit-specific candidate generation and verification to improve illumination and leaf edits while preserving plant structures. Additionally, task-specific evaluation metrics are needed to assess biological correctness, particularly for automatic leaf-count verification, while reducing prompt preparation costs remains important for practical deployment. These efforts can advance reliable agricultural image editing and support data augmentation for agricultural plant recognition and phenotyping.



\clearpage

\section*{LLM Usage Statement}
The authors used ChatGPT (OpenAI) to polish the manuscript's language, improve its overall presentation, and refine the figures. The authors remain fully responsible for all content presented in the paper.

{
\small
\bibliographystyle{ieeenat_fullname}
\bibliography{SearchChord}
}

\end{document}